\documentclass[runningheads, orivec]{llncs}

\usepackage[T1]{fontenc} 
\usepackage{graphicx}
\usepackage{nccmath}
\usepackage{marvosym} 
\usepackage{color}
\usepackage{hyperref}
\usepackage{algorithm}
\usepackage{algorithmic}
\usepackage[utf8]{inputenc}
\usepackage{csquotes}            
\usepackage{lipsum}                 
\usepackage{adjustbox}
\usepackage{siunitx}
\usepackage{float}
\usepackage{listings}
\usepackage{graphbox}
\usepackage{multirow}
\usepackage{booktabs} 
\usepackage{verbatim}
\usepackage[T1]{fontenc}
\usepackage[english]{babel}

\begin{document}
\title{A Simple and Effective Baseline for Attentional Generative Adversarial Networks}
\titlerunning{A Simple and Effective Baseline for Attentional GAN}

\author{
Mingyu Jin\inst{1*}\and 
Chong Zhang\inst{3*}\and
Qinkai Yu\inst{2} \and
Haochen Xue\inst{3} \and
Xiaobo Jin\inst{3}\textsuperscript{\Letter} \and
Xi Yang\inst{3}\textsuperscript{\Letter}
}
\vspace{-10pt}
\authorrunning{C. Zhang, M. Jin, Q. Yu, H. Xue, et al.}
\institute{Electrical and Computer Engineering, Northwestern University, Evanston, U.S.A.
\and
Department of Mathematics Science, University of Liverpool, Liverpool, U.K.
\and 
School of Advanced Technology, Xi'an Jiaotong-Liverpool University, Suzhou, China 
\\
\email{u9o2n2@u.northwestern.edu} \quad \email{sgqyu9@liverpool.ac.uk}  \\
\email{\{Chong.zhang19, Haochen.Xue20\}@student.xjtlu.edu.cn} \\
\email{\{Xiaobo.Jin, Xi.Yang\}@xjtlu.edu.cn} 
\footnote{* Equal contribution. \\  \textsuperscript{\Letter} Corresponding author.}
}

\maketitle  
\vspace{-15pt}

\begin{abstract}
Synthesising a text-to-image model of high-quality images by guiding the generative model through the Text description is an innovative and challenging task. In recent years, AttnGAN based on the Attention mechanism to guide GAN training has been proposed, SD-GAN, which adopts a self-distillation technique to improve the performance of the generator and the quality of image generation, and Stack-GANs, which gradually improves the details and quality of the image by stacking multiple generators and discriminators. However, this series of improvements to GAN all have redundancy to a certain extent, which affects the generation performance and complexity to a certain extent. We use the popular simple and effective idea (1) to remove redundancy structure and improve the backbone network of AttnGAN. (2) to integrate and reconstruct multiple losses of DAMSM (Deep Attentional Multimodal Similarity Model). Our improvements have significantly improved the model size and training efficiency while ensuring that the model's performance is unchanged and finally proposed our \textbf{SEAttnGAN}. Code is available at \href{https://github.com/jmyissb/SEAttnGAN}{https://github.com/jmyissb/SEAttnGAN}.

\end{abstract}

\vspace{-20pt}
\section{Introduction}

Text-to-image technology, at the intersection of computer vision and natural language processing, converts textual descriptions into visual representations. It finds applications in art creation, virtual scene generation, advertising design, and game development etc. ~\cite{isola2017image} \cite{gatys2015neural} \cite{karras2017progressive} \cite{frolov2021adversarial} \cite{hald2020procedural}. 
Within the field of image generation, an influential framework is the Generative Adversarial Network (GAN). By employing adversarial learning, wherein the generator and discriminator are trained simultaneously, the quality of the generated images progressively improves.
GAN-based text-to-image models have rapidly evolved since the introduction of Text descriptions as supervised signals, demonstrated by GAN-INT-CLS~\cite{reed2016generative}. Notable advancements include the Adversarial What-Where Network (GAWWN) with its scene layout estimator~\cite{reed2016learning},  the Pixel Recurrent Neural Networks (PixelCNN) that model image pixel distributions using CNNs~\cite{van2016conditional}, and the Stacked Generative Adversarial Networks (StackGAN and StackGAN-v2) that enhance image quality through multi-stage models or multiple generators and discriminators~\cite{zhang2017stackgan}\cite{zhang2018stackgan++}. Furthermore, the Semantics Disentangling GAN (SD-GAN) addresses latent space factor confusion with a self-distillation mechanism~\cite{yin2019semantics}.


To improve the Text-to-image model performance, Attentional Generative Adversarial Network (AttnGAN) introduces the attention mechanism and the deep attention multimodal similarity model in the generation work, which enables AttnGAN to capture the word-level fine-grained information in the sentence. Thus allowing it to synthesise the advantages of fine-grained details of different subregions of an image by focusing on related words in the natural language description. However, the disadvantages of AttnGAN are also apparent. The AttnGAN contains multiple Attention models and multiple groups of baseline networks, which makes its model structure complex and has many parameters, which may lead to high computational complexity. To solve the problems,

\begin{itemize}
  \item We propse our optimized simplified version, called Simple and effective Attentional Generative Adversarial Networks (SEAttnGAN ). Our work improves the structure of AttnGAN and optimises the number of its parameters, and its performance maintains the same level as before optimisation. 
  \item Additionally, to achieve the corresponding effect even after model simplification and parameter reduction, we introduced semantics in the model. Our model can use the comparative loss of semantics as auxiliary information that can guide the GAN generator to more accurately generate images that match the text description. 
\end{itemize}
\par To solve the large and complex problem of the AttenGAN model, our work proposes a simplified AttenGAN that performs similarly to the original AttenGAN with significantly reduced structure and reduced parameter count. Even in the CUB dataset generation test, we use one-tenth of the number of parameters of AttnGAN (our SEAttnGAN, 26.37M; AttnGAN, 230M) surpasses the image generation quality of AttnGAN, and the time consumption is much less than AttnGAN, which powerfully achieves our motivation.

\par The overall paper is organized as follows: in Section 2, we introduce previous related work on text-to-image task. Section 3 gives a detailed introduction to our method and baseline; in Section 4, our algorithm will be compared with other algorithms qualitatively and quantitatively. Subsequently, we summarize our algorithm and possible future research directions.
\vspace{-10pt}

\section{Related Work} 

GANs~\cite{goodfellow2020generative}  have the remarkable capability to generate realistic images that are indistinguishable from real ones.
Then Dong et al. proposed a method to generate realistic images using only natural language written descriptions, where they changed the structure of GAN by training a style encoder network to invert the generator network  \cite{dong2017semantic}.

While current methods can generate images with novel properties, they do not preserve the text-independent details of the original image. To address this issue, Nam et al. propose TAGAN (Text Adaptive Generative Adversarial Networks) to generate semantically manipulated images while preserving text-independent details \cite{nam2018text}. The key idea of PPGN (Plug \& Play Generative Networks)\cite{nguyen2017plug} is to feed random noise vectors into generative models and combine them with optimization techniques to guide the generative process. LAPGAN (Laplacian Generative Adversarial Network)\cite{denton2015deep} is a generative model framework combining Laplacian pyramid decomposition and GAN, aiming to generate high-resolution images from coarse to fine in a multi-scale progressive manner. MirrorGAN \cite{qiao2019mirrorgan} achieves text-to-image conversion by learning fine-grained correspondences, in which reverse information propagation during generation can promote consistency between text and images, and in the process of reverse generation through The discriminator provides an additional supervisory signal. In text-to-image synthesis, conditional GANs have been widely used, where a generator will take a text description as input and generate an image \cite{mirza2014conditional} that matches the description. Recent studies have shown that tuning GANs based on different factors such as object attributes, styles, and layouts can improve the quality and diversity of generated images.

AttnGAN is a symbolic attention-based \cite{xu2018attngan} that improves the quality and relevance of generated images by focusing on important parts of textual descriptions. These mechanisms have been used to guide generators to notice connections between words and phrases and use them to create corresponding images. A multimodal GAN \cite{amirian2019social} has been developed to generate multiple images corresponding to a single textual description, and these diverse high-quality images can provide different interpretations for the same textual description. VAE(Variational Autoencoder) is a type of generative model that has been explored for text-to-image synthesis \cite{kingma2013auto}. In VAE-based text-to-image synthesis, the model learns a latent space representation of the input text descriptions, which can be used to generate images. One of the primary advantages of using VAE-based models is their interpretability. But, there is still room for improvement to generate photo-realistic images that match the input textual description. PixelCNN was a generative model generates images pixel by pixel, where each pixel is conditioned on the previous pixels \cite{van2016conditional}. The key idea behind PixelCNN is to use a convolutional neural network (CNN) to model the distribution of the image pixels. PixelCNN has been shown to be effective at generating high-quality images, it has also been used for text-to-image synthesis, where the image is generated conditioned on an input text description.

GAN-INT-CLS is the first application study that adds textual descriptions (i.e., sentence embedding vectors) as supervised signals to image generation \cite{reed2016generative}. The main goal of GAWWN is to estimate the scene layout of an image while generating it \cite{reed2016learning}. GAWWN improves the text-to-image performance by introducing Scene Layout Estimator in the image generation process \cite{zhang2017stackgan}. StackGAN (Stacked Generative Adversarial Networks) is used to gradually generate high-resolution images. This paper constructs a multi-stage generative model, which aims to solve the challenge of traditional GAN models in generating detail-rich images. Stackgan-v2 This paper improves and extends StackGAN by adding multiple generator and discriminator networks \cite{zhang2018stackgan++}, introducing a super-resolution network, and introducing a triangular loss function and a classification loss function. SD-GAN mainly solves the confounding problem of the traditional GAN model in representing factors in the latent space and introduces the mechanism of self-disentanglement learning to train the generator and discriminator networks \cite{yin2019semantics}. Training of the process-aided generator for "self-distillation" by augmenting the Distiller. SD-GAN mainly solves the confounding problem of the traditional GAN model in representing factors in the latent space\cite{ma2019sd}, and introduces the mechanism of self-disentanglement learning to train the generator network and the discriminator network \cite{zhu2019dm}. Training of the process-aided generator for "self-distillation" by augmenting Distiller. 
\vspace{-10pt}

\section{Method}

\begin{figure}
    \centering
    \includegraphics[width=1\textwidth]{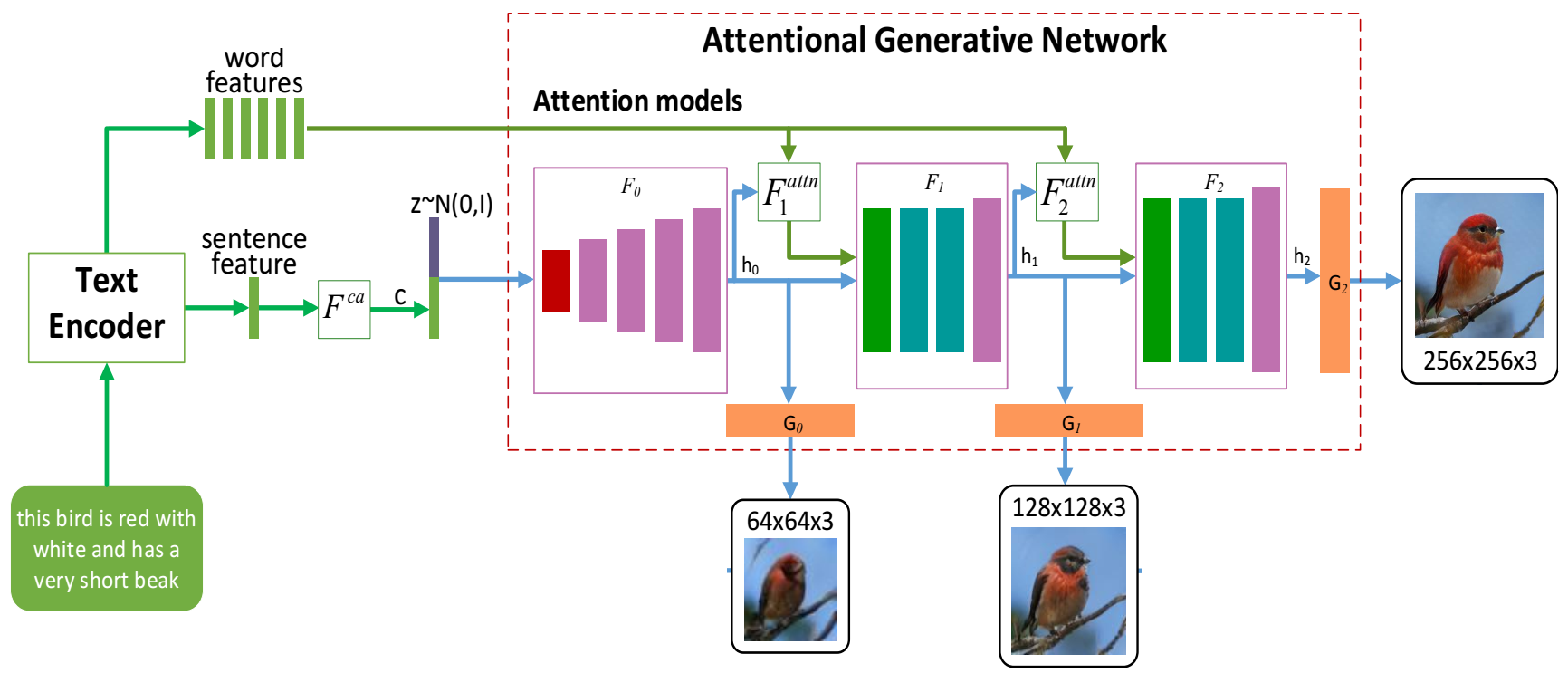}
    \caption{Overview architecture of AttnGAN}
    \label{attn}
\end{figure}

\subsection{Background: AttnGAN}

AttnGAN \cite{xu2018attngan} achieves fine-grained text-to-image generation through attention-driven multi-stage refinement, which focuses on related words to synthesize some details in different regions of the image. 

The AttnGAN model consists of $m$ generators ($G_{0}, G_{1},..., G_{m-1}$), each generator will generate an image as follows

\begin{eqnarray}
     H_0 & = & F_0(z,F^{ca}(T)) \\
     h_i & = & F_i(h_{i - 1}, F^{attn}(T,H_{i - 1})) \\
     \hat{x}_i & = & G_i(h_i)    
\end{eqnarray}
where $i = 1,2,\cdots,m - 1$, $z$ is a noise variable following a Gaussian distribution, $T$ is a matrix of word vector representations, and $F^{ca}$ uses conditional augmentation to encode each word vector in $T$. The attention network $F^{attn}$ will generate a weighted representation of the word vector according to the relationship between the image vectors and the text vectors.

The attn network $F^{attn}$ takes word features and image features as input: 1) maps word features to image feature space; 2) generates a new representation of word features according to the correlation between word features and image features, which is a linear weighting of image features combination. The specific description is as follows
$\hat{T}$  =  $U T$ \\
\begin{equation}
    \alpha_{i,j}  = \frac{\exp (\hat{t}_i^T h_j)}{\sum_{k = 1}^{m} \exp (\hat{t}_i^T h_k)},\quad
c_i  =  \sum_{j = 1}^m \alpha_{ij} h_j,
\end{equation}
where $U$ is the transformation matrix, which transforms word vectors $t_i$ into the semantic space of image features $h_i$.

AttnGAN model jointly optimizes generative adversarial loss and deep attention multimodal similarity loss. The generative adversarial loss contains conditional distribution and unconditional distribution likelihood function loss, where the unconditional distribution likelihood loss determines whether an image is real or fake, and the conditional distribution likelihood loss determines whether an image and a sentence match. DAMSM learns two neural networks to map image sub-regions and sentence words into a common semantic space respectively to compute fine-grained loss for image generation.

\subsection{Our Proposed Method}

In AttnGAN, multiple generators generate images of increasing scale and use an attention mechanism for each generator. In order to reduce the complexity of the model, we replace multiple generators with upsampling modules to output ever-increasing feature maps, and use attention only on feature maps at one scale, and to align with word embedding vectors, we insert multiple downsampling modules, as shown below in Fig. \ref{method}.

In this paper, we simplified the network structure of AttnGAN. For the input of the generator, we mix sentence features with noise. After data augmentation, the input is first reshaped through a fully connected layer. Then apply a series of Up Block layers to extract image features. The Up-Block layer is composed of an upsample layer, a redisual block, and DF-Blocks to fuse the text and image feature during the generation process. The image features obtained are used as  input and enter the attention layer together with the text features. Finally, the image features are re-extracted by a layer of Up-Block and converted to image features by a convolution layer. The detail architecrure shown in Figure \ref{method}

\begin{figure}[htbp]
    \includegraphics[width=1.0\textwidth]{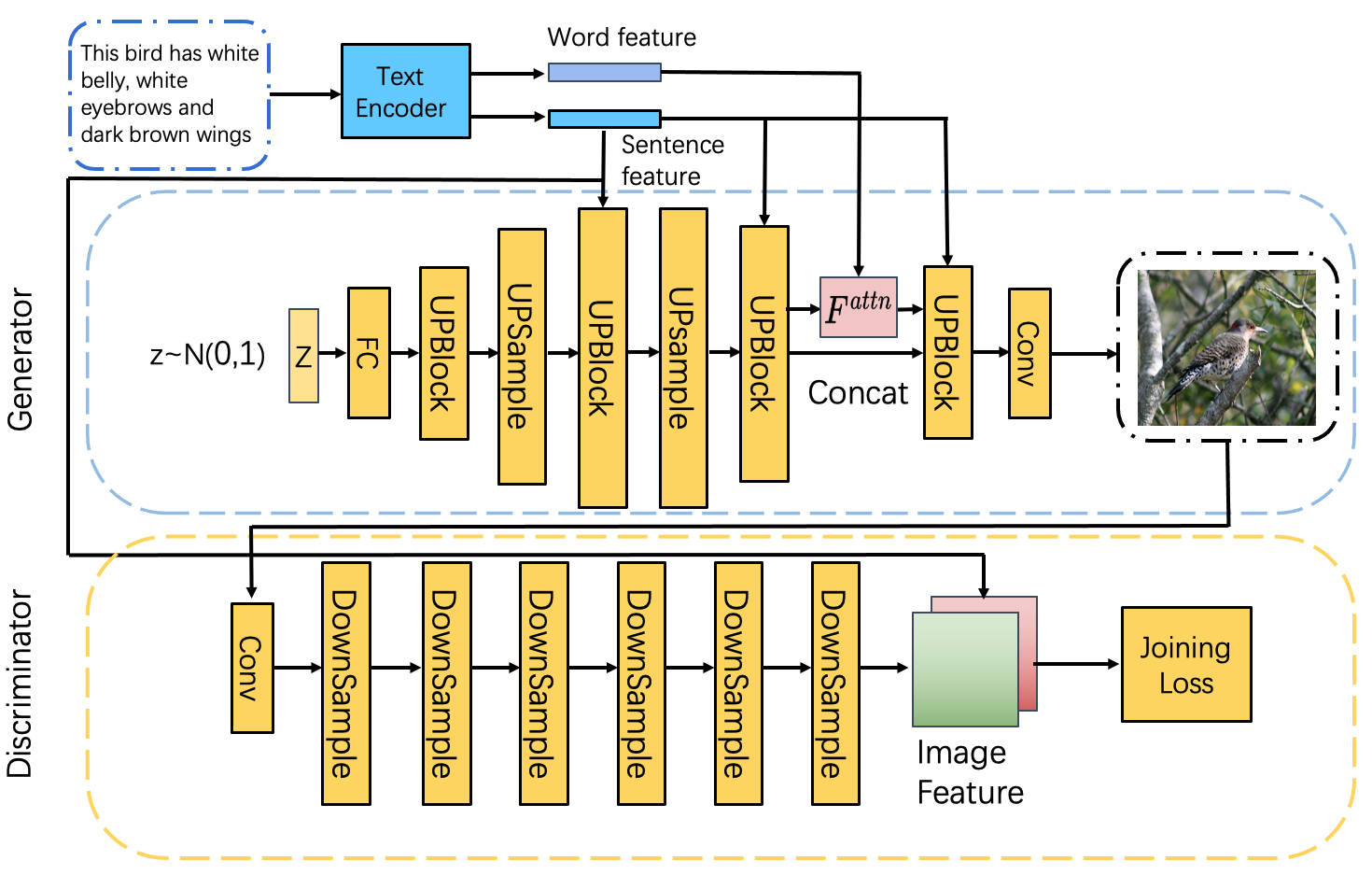}
  \caption{Architecture of SEAttnGAN. FC: fully connected layer. UPBlock: residual block + DFBlock. DownBlock: downsample + residual block}
  \label{method}
\vspace{-20pt}
\end{figure}
\begin{figure}[htbp]
  \centering
  \begin{minipage}[b]{0.4\textwidth}
    \centering
    \includegraphics[width=1.2\textwidth]{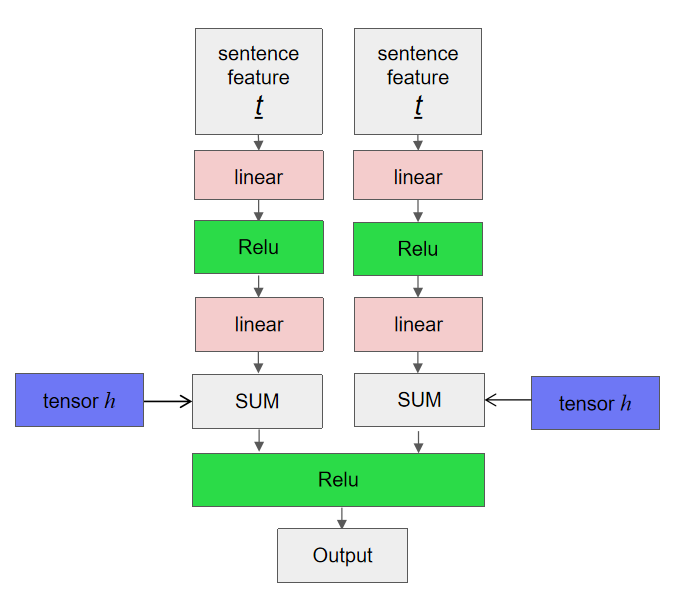}
    \label{dfblock}
  \end{minipage}
  \hfill
  \begin{minipage}[b]{0.4\textwidth}
    \centering
    \includegraphics[width=\textwidth]{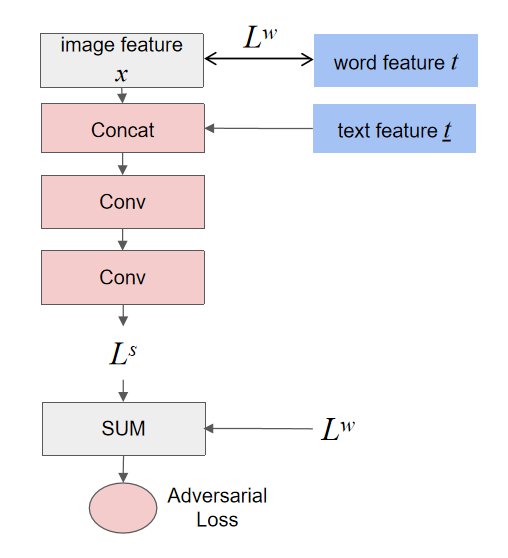}
    \label{loss}
  \end{minipage}
  \caption{(1)The architecture of the DFblock: 2 layers linear mlp, tensor from the previous layers. (2)The architecrure of our model's loss function}
  \label{DFblock and loss}
\end{figure}

We simplified the existing DFblock structure and used a new DFblock structure consisting of a two-layer linear mlp structure and an additional hidden layer (with ReLU activation) to further extract sentence features, map sentence features from 256 dimensional space to 64 dimensional space, and obtain new representation features. The new structure has the characteristic of lightweight, reducing the number of model parameters, simplifying the calculation process, and accelerating training speed. The detail architecrure shown in Figure\ref{dfblock}\\
The discriminator converts images into features through a series of DownBlock. Then the sentence vector will be replicated and concatenated with image features. The discriminator distinguishing generated images from real samples, the discriminator promotes the synthesize images with higher quality and text-image semantic consistency. And the word feature is used at word level to compute loss by measures the similarity with image-text.\\
The text encoder is a bi-directional Long Short-Term Memory(LSTM). and we directly use the pre-trained model provided by AttnGAN.\\
The image encoder is a Convolutional Neural Network(CNN) that maps images to semantic vectors. we directly use the Inception-v3 model pretrained on ImageNet.
To gnerate realistic images with multiple levels(i.e.,sentence level and word level) of conditions,the loss function of our network is defined as:
\begin{equation}
    \mathcal{L} =\mathcal{L}_{G}+ \gamma \mathcal{L}_{D}
\end{equation}
where
\begin{equation}
\begin{aligned}
        &\mathcal{L} _{G} = -E_{G(z)}\sim\ p_{g}[D(G(z)),t]\\
        &\mathcal{L}_{D} = \lambda \mathcal{L}^{s}+(1-\lambda) \mathcal{L}^{w}.
\end{aligned}
\end{equation}
Here $\gamma$  and $\lambda$ is hyperparameter to balance the terms of equation. 
Here we defined those two loss function, respectively.
\begin{equation}
\begin{aligned}
\mathcal{L}^{s}= & - E _{x \sim P _r}[\min (0,-1+D(x, \underline{t}))] \\
& -(1 / 2) E _{G(z) \sim P _g}[\min (0,-1-D(G(z), \underline{t}))] \\
& -(1 / 2) E _{x \sim P _{m i s}}[\min (0,-1-D(x, \underline{t}))] \\
& +k E _{x \sim P _r}\left[\left(\left\|\nabla_x D(x, \underline{t})\right\|+\left\|\nabla_t D(x,\underline{t})\right\|\right)^p\right] \\
\end{aligned}
\end{equation}
where $z$ is the noise vector; $\bar e$ is the sentence vector; $P_{g},P_{r},P_{mis}$ represent the synthetic data distribution, real data distribution, and mismatching data distribution, respectively.
\begin{equation}
    \begin{aligned}
        \mathcal{L}^{w} = -\sum^{M}_{i=1} log \frac{exp(\mu \cdot sim(c_{i},t_{i}))}{\sum^{M}_{j=1}exp(\mu  \cdot sim(c_{i},t_{j}))}
        \end{aligned}
\end{equation}
where
\begin{equation}
    c_{i}= \sum^{n-1}_{j=0} \alpha_{j}x_{j}, \quad \alpha_{j}=\frac{exp(\mu_{1}s_{i,j})}{\sum^{n-1}_{k=0}exp(\mu_{1}s_{i,k})}
\end{equation}
$s$ is a matrix product of entire image feature and word feature defined by $t^{T}x$. 
$\mu$ and $\mu_{1}$ is a smoothing factor determined by experiments. In this batch of sentences, only $t_{i}$ matches the image $x_{i}$, and treat all other $M-1$ sentences as mismatching descriptions. $sim(c_{i},t_{i})$ is the cosine similarity between $c_{i}$ and $t_{i}$ which equal to $\frac{c_{i}^{T}t_{i}}{\mid \mid c_{i}\mid \mid \mid \mid t_{i}\mid \mid  } $. Based on experiments on validation set, we set the hyperparameters in this section as: $\gamma=5$, $\lambda=0.2$, $\mu=5$, $\mu_{1}=10$. The detail architecrure shown in figure \ref{loss}
\vspace{-10pt}

\section{Experiment}
\textbf{Datasets.} We selected two datasets: CUB Bird \cite{WahCUB_200_2011} and COCO \cite{lin2014microsoft}. The CUB dataset contains 11,788 bird images of 200 species, each with descriptions in 10 languages. Meanwhile, the COCO dataset has 80,000 images for training and 40,000 images for testing, each with descriptions in five languages.

\noindent \textbf{Experiment settings.} Both our SEAttnGAN generator and discriminator use the Adam optimizer. The generator uses the parameters $\beta_{1} = 0$, $\beta_{2} = 0.9$ and the learning rate is $0.0001$ \cite{kingma2014adam}; while the discriminator uses the parameters $\beta_{1} = 0.0$, $\beta_{2} = 0.9$, the learning rate is $0.0004$. 

\noindent \textbf{Evaluation metrics.} Finally, we choose IS (Inception Score) and FID (Fréchet Inception Distance) to evaluate the quality of the generated image. IS measures \cite{salimans2016improved} the KL divergence between the generated image category distribution obtained by a pre-trained Inception V3 model and the real image category to represent the authenticity of the generated image. The smaller the KL divergence, the closer the category distribution of the generated image is to the real image. Higher IS values indicate better diversity and authenticity of generated images. It's not perfect though, as it may not capture the quality of detail and realism of the resulting image. Therefore, when evaluating image generation models, it is usually necessary to combine other indicators and human subjective evaluation for comprehensive consideration.

Frechet Inception Distance (FID) \cite{heusel2017gans} is a metric used to measure the quality of image generation. It is based on the statistical feature difference between the generated image and the real image: it first uses a pre-trained image classifier (such as the Inception V3 model) to extract the feature representation of the real image and the generated image, and then uses the Fréchet distance to measure the difference between the two feature distributions. The difference between the sample mean and the covariance matrix of. The lower the value of FID, the better the quality of the generated image, and the higher the coincidence with the real image distribution. The advantage of FID over other metrics is that it can capture more subtle differences in image quality and has a high correlation with human subjective evaluation results. However, FID also has some limitations, such as requiring a pre-trained classifier and a large number of image samples.

\vspace{-10pt}
\subsection{Quantitative Comparison}

In this section, we compare our new method with baseline methods and some other methods, here we report quantitative metrics such as IS and FID on CUB Bird\cite{WahCUB_200_2011} and COCO datasets\cite{lin2014microsoft}. In particular, we also compare the memory and time consumed by these methods during inference. During model training, some models use additional pre-training models: SD-GAN\cite{yin2019semantics} uses third-party trained COCO pre-training models\cite{rombach2021highresolution}, CPGAN\cite{liang2020cpgan} uses additional pre-training YOLO v3\cite{farhadi2018yolov3}, XMC-GAN\cite{zhang2021cross} use additional pre-trained VGG-19\cite{simonyan2014very} and Bert\cite{devlin2018bert}, DAE-GAN use additional NLTK POS markers and design manually Rules and TIME use an additional two-dimensional positional encoding.

\begin{table}[htbp]
        \caption{The results of FID, IS and NOP comparison on dataset CUB bird and COCO.}
	\begin{center}
	    \begin{tabular}{l|c|c|c|c|c}
		\hline {Model}  & \multicolumn{2}{|c|}{ CUB } & \multicolumn{1}{|c|}{ COCO } & \multicolumn{2}{|c}{Cost}\\
		\cline { 2 - 6 } & IS $\uparrow$ & FID $\downarrow$ & FID $\downarrow$ & Memory $\downarrow$ & Inference Time $\downarrow$ \\
		\hline StackGAN\cite{zhang2017stackgan}  & 3.70 & - & - & - & - \\
		StackGAN++\cite{zhang2018stackgan++}& 3.84 & - & - & - & -\\
		MirrorGAN\cite{qiao2019mirrorgan}  & 4.56 & 18.34 & 34.71 & - & - \\
		SD-GAN\cite{yin2019semantics}  & 4.67 & - & - & 335M & 6.18s\\
		DM-GAN\cite{zhu2019dm}  & 4.75 & 16.09 & 32.64 & 46M & 7.06s\\
		CPGAN\cite{liang2020cpgan}  & - & - & 55.80 & 318M & -\\
		TIME\cite{DBLP:conf/aaai/LiuSZME21} & \textbf{4.91} & \textbf{14.30} & 31.14 & 120M & -\\
		XMC-GAN\cite{zhang2021cross} & - & - & \textbf{9.30} & 166M & -\\
            DF-GAN\cite{tao2020df} & 4.46 & 18.23 & 41.83 & 46.79M & 3.71s \\
		DAE-GAN\cite{ruan2021dae}  & 4.42 & 15.19 & 28.12 & 98M & -\\
            \hline AttnGAN \cite{xu2018attngan} & 4.36 & 23.98 & 35.49 & 230M & 4.05s\\
		\textbf{SEAttnGAN(ours)} & 4.32 & 15.03 & 34.78 & \textbf{26.37M} & \textbf{0.11s} \\
		\hline
	    \end{tabular}
            \vspace{-20pt}
	\end{center}
    \label{tab1}
\end{table}

As shown in the results in Table \ref{tab1}, our method achieves comparable performance to other methods on IS and FID metrics for both datasets. In particular, our method achieves better performance compared to the baseline model AttnGAN, for example, on CUB, our method achieves a FID value of 15.03, which is better than AttnGAN's 23.98. In terms of inference time, our method runs about 45 times faster than the baseline method (from 4.05 seconds to 0.11 seconds). Considering the memory footprint of the model, our method is much lower than other methods. Compared with the baseline method AttnGAN, it reduces the memory footprint to about 1/9 (26.37M vs. 230M). Our SEAttnGAN did not achieve an absolute advantage in COCO, which may be due to the difficulty of training generative models in a short period of time due to the large variety of items in the COCO dataset. Comparing the FID values on the two data sets of CUB and COCO, we also found that the image generation task on COCO is more difficult than that of CUB. DF-GAN\cite{tao2020df} using the same simple and effective ideas also significantly reduce model size. Still, our SEAttnGAN outperforms DF-GAN in generating image quality (IS, FID) and generation time. Besides, compare with the original loss in DF-GAN, our new loss saw an outstanding improvement. The detail is showed in figure \ref{tab2}
\vspace{-10pt}
\begin{table}[htbp]
        \caption{The results of FID, IS comparison on dataset CUB bird}
	\begin{center}
	    \begin{tabular}{|c|c|c|}
		\hline {Model}  & \multicolumn{2}{|c|}{ CUB } \\
		\cline {2-3} & IS $\uparrow$ & FID $\downarrow$\\
		\hline SEAttnGAN with original loss & 2.88 & 34.71  \\
		  \hline SEAttnGAN with new loss & 4.29 & 15.64 \\
            \hline
	    \end{tabular}
            \vspace{-20pt}
	\end{center}
    \label{tab2}
\end{table}

\vspace{-10pt}
\subsection{Qualitative Comparison}
\vspace{-5pt}
\begin{figure}[ht]  
    \includegraphics[width=1.0\textwidth]{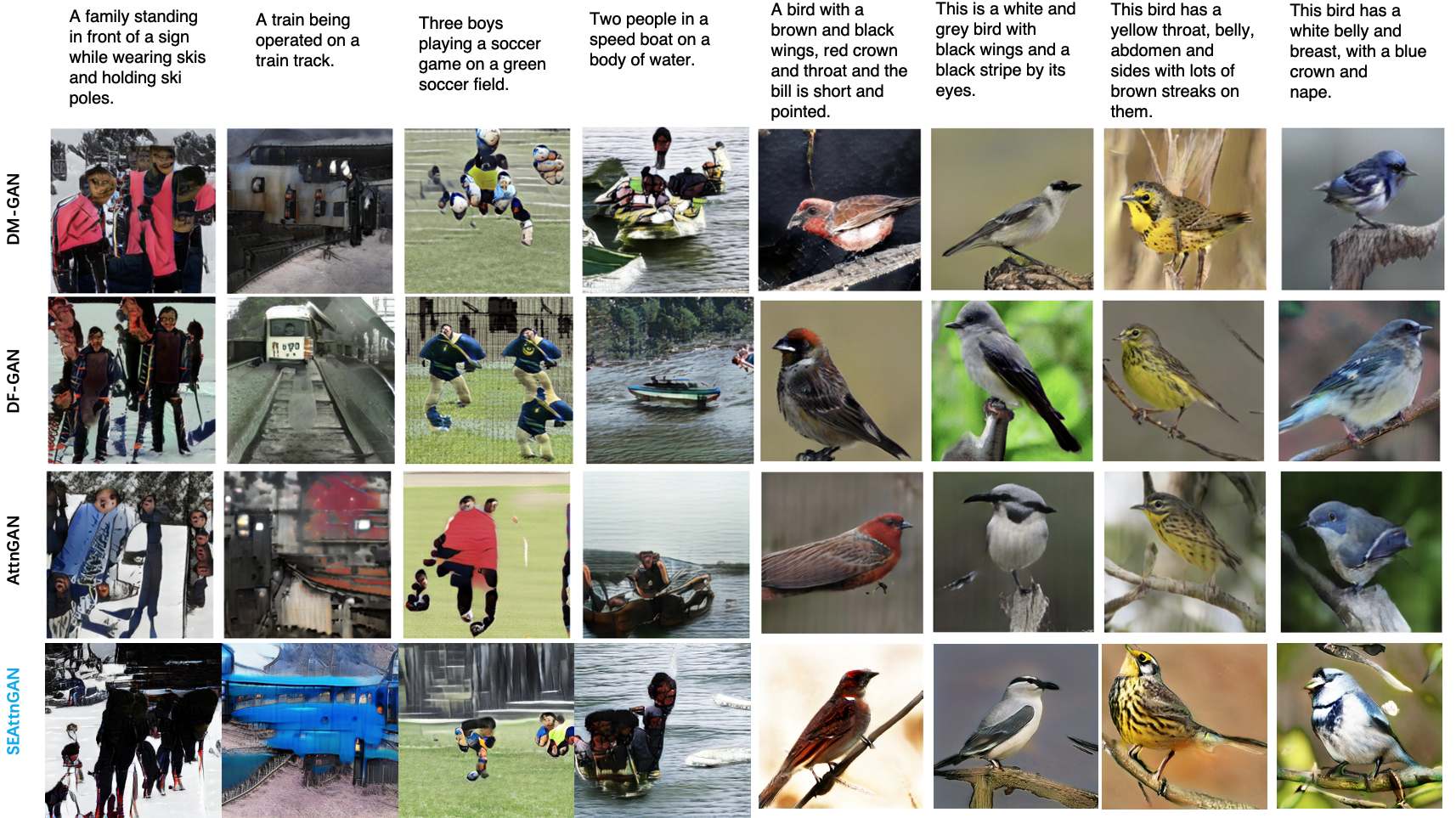}
    \vspace{-10pt}
    \caption{The architecture of the SEAttnGAN. FC: fully connected layer. UPBlock: upsample + residual block + DFBlock. DownBlock: downsample + residual block}
      \label{fig:gan-results}
\vspace{-10pt}
\end{figure}

We compare the visualization of the model with three different models in Fig. \ref{fig:gan-results}: AttnGAN \cite{xu2018attngan}, DF-GAN and DM-GAN \cite{zhu2019dm}. As can be seen from Fig. \ref{fig:gan-results}, the results obtained from AttnGAN and DF-GAN are not as realistic as those obtained from DM-GAN and our model. The difference between the bird and the background is less obvious, especially in the leg and tail areas. Also, the feathers on the bird's back lack a certain clarity. The image generated by DM-GAN also loses part of the image. In contrast, the images generated by our model exhibit a striking feature: a high degree of discrimination between the bird and the background. The pattern of the bird in our model is easy to discern. In addition, the background in the images of our model has stronger brightness than those generated by the other three models.

Finally, we also observe from Fig. \ref{fig:gan-results} that the images in the four columns on the left are distorted on all three models on the COCO dataset, which means that some images are missing, duplicated, or folded. However, compared to images generated by the other three models, the background in our model's images shows stronger brightness, adding to the visual appeal and overall aesthetics of the output.

\vspace{-10pt}
\section{Conclusion}
\vspace{-10pt}
Based on the results above, it appears that AttnGAN utilizes two attention models during the fine-grained image generation process, which seems unnecessary. However, after implementing a simple and effective target improvement, the model's size was significantly reduced without any significant decrease in performance indicators like FID and IS when compared to the original AttnGAN version. Even in the CUB bird's FID item, we beat our improved object AttnGAN with one-tenth of the model size, which is an exciting result. Our improvement involved eliminating one set of attention models and adding the upsample and residual block, resulting in better image quality. Additionally, our combination of loss functions for the three stages of SEAttnGAN has proven to be effective.
\vspace{-10pt}
\section{Future work}
\vspace{-8pt}
The limitation of our model mainly reflects on the low performance of the COCO dataset. It occasionally generates defective images, meaning it needs to run the model several more times to generate the correct image. Future work will focus on taking methods to overcome these shortcomings. Initially, the number of parameters in our model will be minimised to render our model simple and effective. We anticipate improved performance and more reliable image generation by reducing the complexity. our attention will be directed towards enhancing the Attention-GAN model itself. We will explore and implement novel techniques to refine and optimise its architecture, enhancing its ability to generate high-quality images. Eventually, the Simple and effective philosophy can be extended to most models that are large but widely used by industry.
\vspace{-10pt}
\section{Authorship Statement}
\vspace{-8pt}
All authors contributed to the study's conception and design. Mingyu Jin and Chong Zhang completed the construction and debugging of the basic network framework of SEAttnGAN. Simple and effective improvements to AttnGAN ideas and improvements to the Attention model were completed by Chong Zhang. Loss function reconstruction and SEAttnGAN Mathematical interpretation by Qinkai Yu. Optimising the Attention Model combined with semantic information was done by Haochen Xue. All authors wrote the manuscript, and all authors commented on previous versions. Xiaobo Jin supervised this work and made a comprehensive revision and reconstruction of the work. All authors read and approved the final manuscript.

\bibliographystyle{plain}
\bibliography{main}

\end{document}